# Optimal Categorical Attribute Transformation for Granularity Change in Relational Databases for Binary Decision Problems in Educational Data Mining

Paulo J.L. Adeodato
Centro de Informática
Universidade Federal de Pernambuco
Recife-PE, Brazil
pjla@cin.ufpe.br

Fábio C. Pereira
Faculdade de Odontologia
Universidade de Pernambuco
Camaragibe-PE, Brazil

Rosalvo F. Oliveira Neto
Engenharia da Computação
Universidade Federal do Vale do São Francisco
Juazeiro-BA, Brazil
rosalvo.oliveira@univasf.edu.br

*Abstract*— This paper presents an approach for transforming data granularity in hierarchical databases for binary decision problems by applying regression to categorical attributes at the lower grain levels. Attributes from a lower hierarchy entity in the relational database have their information content optimized through regression on the categories´ histogram trained on a small exclusive labelled sample, instead of the usual mode category of the distribution. The paper validates the approach on a binary decision task for assessing the quality of secondary schools focusing on how logistic regression transforms the students´ and teachers´ attributes into school attributes. Experiments were carried out on Brazilian schools´ public datasets via 10-fold cross-validation comparison of the ranking score produced also by logistic regression. The proposed approach achieved higher performance than the usual distribution mode transformation and equal to the expert weighing approach measured by the maximum Kolmogorov-Smirnov distance and the area under the ROC curve at 0.01 significance level.

**ACM-Classification**
I.2 ARTIFICIAL INTELLIGENCE
H.2.8 Database Applications, J.1 ADMINISTRATIVE DATA PROCESSING-Education;

## I. Introduction

One of the most important aspects in decision support systems (DSS) is to explore all data and information for improving their performance while preserving the domain semantics on the attributes. A difficulty arises when the databases contain information in several granularities different from that of the decision level. This occurs in relational databases where there are *1:n* relationships among the entities defined by their tables in various hierarchical levels.

Relational Data Mining (RDM) [1] is a research area in the borderline between databases and artificial intelligence that handles information in different granularities. Its "proposicionalization" approach produces a denormalized table in the decision grain [1], as required by most knowledge extraction algorithms. That forces the summarization of lower level attributes´ distributions in one or few indicators/features on the decision level.

Fig. 1 presents the scenario for a domain such as education, with schools being assessed by their own and their students´ and their teachers´ features in a hierarchical structure. "How to assign the students´ father education level attribute as a school attribute based on the distribution of its students?" is an example of question that illustrates the point.

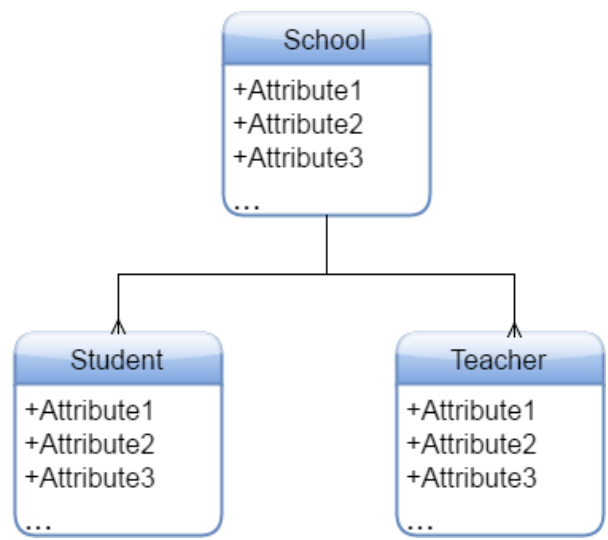

Fig. 1. Schema database representation of schools

A further constraint to the granularity concept adopted here is that all lower level grains do not contain objects that form a series or have any type of order relation. That is different from the sequences of events along the temporal dimension considered in RFM analysis (Recency, Frequency and Monetary value analysis) applied in behavior scoring for credit risk analysis [2].

In such context, this paper proposes a granularity transformation that uses regression techniques to optimize the information content of single categorical attributes. The transformation maximizes the information gain towards the target variable in binary decision problems, preserving the conceptual value of the attribute in the original lower grain. It represents an improvement of a previous approach, which does a similar transformation based on human expert´s knowledge if that is available [3]. The proposal is tested on the school quality assessment problem based on their teachers and students' data.

This paper is organized in 5 more sections. Section II presents an overview the topic on related fields and discusses how they deal with the granularity change problem. Section III presents the proposed optimization approach. Section IV describes the experimental project to validate the proposed approach detailing each step from database selection to the performance assessment metrics. Section V presents the results and discusses its impacts. Section VI concludes the paper summarizing the main contributions of the proposed approach, analyzing its limitations and suggesting potential ways for expanding it to other types of attributes and supervised problems.

## II. Related Fields of Research

Granularity change has been an important topic in several research areas such as Relational Data Mining (RDM) [1], Granular Computing (GrC) [4] and Function Concept Analysis (FCA) [5] with different approaches, but partial overlap on their objectives towards how to preserve the relevant concepts in the transformation. These areas all deal with high complexity data mining problems in the borderline between databases and artificial intelligence.

A key aspect in many high complexity data mining problems is the proper understanding and transformation of the relationships among the entities defined by the tables of relational databases. Several areas have proposed different approaches for dealing with the matter from different perspectives and levels of formal background.

The database structure associated with the decision-making process workflow defined by the project goals determine the so called grains of information of the problem to be solved. In statistics, the lowest hierarchical level of the database structure (finest granularity) stores the so-called microdata statistics [7].

### A. Relational Data Mining (RDM)

Relational Data Mining (RDM) [1] refers to hierarchies, entities, granularities in relational databases for knowledge extraction. For the present context, in most cases, the approaches are focused on aggregation for clustering or discovery of association rules applications [1] which are based on the non-supervised learning paradigm. Guo and Viktor [6] have proposed an improvement for constructing artificial neural networks classifiers starting with an aggregation stage as well. In general, the RDM approach aims at forming higher level hierarchical concepts by collecting lower level attributes related to them [1,6]. These approaches propose automatic transformations directly from a database without taking into account application other domain characteristics, different from the transformations to capture temporal relations in credit risk analysis [2].

Aggregation can be seen as an operation that summarizes a set of values related to variables; a distribution. In the case of numerical variables, descriptive statistics can be used, such as the minimum value, the mean, the median, the maximum value are the first recommendation. For categorical variables, the typical option is the distribution mode (the most frequent value) [7].

The typical way to represent a distribution by a single or few concepts in the decision grain is done by the taking the lowest order momenta of the probability distributions. That idea is present in the moment invariant approaches used for feature extraction in image processing [8], for example. With the exception of anomaly detection such as fraud detection [9], cancer diagnosis [10] or pulmonary embolism diagnosis [11] which uses the distributions´ extreme values, most applications use the central tendency of the distribution as decision level feature.

### B. Granular Computing (GrC)

Granular Computing (GrC) [4] is another area of research which also deals with the change of granularity in relational databases. It forms higher level hierarchical concepts by collecting lower level attributes focusing either on clustering or discovery of association rules applications [4] thus based on non-supervised learning. However granular computing applies mainly rough set theory and fuzzy logic as tools for achieving its goals. It must be emphasized that it differs from the approach proposed here in two main aspects: no supervised optimization and no embedding of concept from a single lower level attribute. In general terms, granular computing can be considered as a field of multidisciplinary study, dealing with theories, methodologies, techniques and tools that make use of granules in the process of problem solving [12].

### C. Formal Concept Analysis (FCA)

Formal Concept Analysis (FCA) [5] is an approach of very high level of abstraction that, if properly instantiated to very specific conditions, could lead to structures similar to RDM and GrC. It is a branch of applied mathematics and provides a framework called lattice of concepts (based on partial ordering) that presents the relations between objects and values of attributes in diagrammatic format. These structures can be used in data mining tasks [13,14] and make explicit the search space in classification methods. However, the benefits of using conceptual cross-linking are accompanied by costly construction and manipulation [15] in a grid that grows exponentially in relation to the formal context that is composed of objects, attributes, attribute values, and incidence relationships. Furthermore, it is a very broad approach and does not give clear paths on how to optimize the concepts in a univariate supervised learning process.

### D. Weighted Granularity Transform (WGT)

Recently, a new approach has been proposed for granularity change inspired by and tested on education problems. It is an approach based on human expert´s knowledge to make the granularity transformation [3] that this paper will refer to as the Weighted Granularity Transform (WGT). The human expert sets weights for each category of the attribute to express its propensity towards the binary target in a way to produce a monotonic propensity mapping.

For example, for the faculty education distribution of the school, the weights were arbitrarily set as w(PhD)=4, w(MSc)=3, w(MBA)=2 and w(BSc)=1, just preserving an order relation according to the expert's domain knowledge. The attribute transformation was the weighted sum of the relative frequencies of the lower level categories in a single continuous scalar indicator.

That produced a statistically significant improvement in performance compared to the typical distribution mode transformation.

Despite that improvement, that approach presents some drawbacks this paper attempts to solve:

1. It can only be applied when there is semantics on the attribute´s categories,
2. It depends on human expert´s capacity to capture that semantics as knowledge
3. This knowledge has to be expressed at least as an order relation among the attribute´s categories, usually by setting arbitrary linear weights, and
4. This human knowledge is not always confirmed on the data.

### III. Proposed algorithm – Regression Granularity Transform (RGT)

As stated in the Introduction, this paper focuses on transforming relational databases, that represent systems composed of subsystems in various hierarchical levels and the example of school quality assessment will be used to illustrate the ideas put forward here. The concept of "data granularity" (or "information granularity") is embodied in so-called "grain" that represent the relationship-entity levels for the desired data mining solution.

In this work, the main aspects of understanding the granularity relations in a data mining project are:

1. Allow the modeler to identify subsystems and their individual and collective features to enable the generation of subsets of data with statistical independence,
2. Allow the modeler together with the domain human expert identify transformations of high added value for information granularity change and
3. Allow the modeler to embed statistical knowledge from the data on the transformations for granularity change.

All these aspects are taken into account in a Domain-Driven Data Mining (D3M) approach [16] for problem solving.

Different from the approaches presented in Related Fields of Research, the proposed approach is focused on binary decision problems and attempts to optimize the attribute information content towards the target class.

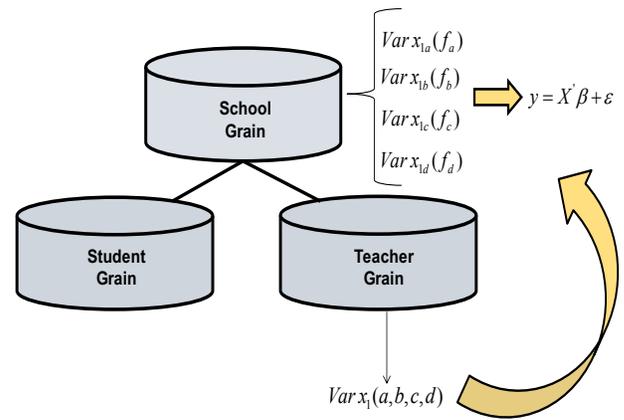

Fig. 2. Proposed approach for granularity transformation.

The optimization is achieved by regression on the categorical attribute distribution having its histogram with the categories relative frequencies as input and the decision level target as output, as depicted in Fig. 2.

The histogram is built on an independent data sample extracted from the modelling dataset used for this sole purpose and discarded afterwards to prevent optimistic bias because of the use of a posteriori information.

Any type of regression can do the job but logistic regression [17] is recommended for its non-linear mapping capacity, ease of interpretation and for its low data consumption compared to other good techniques such as neural networks.

The algorithm can be summarized as follows:

1. Select the categorical attributes with more than 2 categories;
2. Extract a data sample stratified by the target class on the decision grain from the modelling data set;
3. Build the attribute histogram for each example on the decision grain;
4. Run the regression algorithm for parameter estimation (learning);
5. Discard this data sample;
6. Apply the transformation learned through regression to the remaining modelling data sample.

The granularity transformation produces a continuous indicator of the categorical attribute optimized to the target variable on the decision grain still preserving the concept of the original lower grain attribute.

### IV. Experimental Project

The experiments were carried out on the binary problem of assessing the quality of Brazilian private secondary schools [3,18], using logistic regression as classifier and using the Area Under the Receiver Operating Characteristic (ROC) curve and the Maximum Kolmogorov-Smirnov (KS) distance as performance metrics. The goal was to compare if the proposed algorithm would improve the overall classification performance against the previous weighted score algorithm and the usual mode for granularity transformation of the lower level categorical attributes. To verify significant differences, two tailed paired t-test was applied in a stratified 10-fold cross validation procedure at 0.01 significance level [19].

The database consisted of 4,400 schools with microdata from the National Secondary School Exam 2012 (ENEM 2012) and the School Census 2012 [18] with the target class defined as the schools having their students´ average score on the top quartile, as described in Adeodato´s paper [3].

Only the categorical attributes in the lower level grain were considered in the school model. The modelling process was further constrained to the attributes where human expertise could clearly contribute to define the weights. So the only attributes used in the experiments were the Father Education Level and Mother Education Level for the students' database and Teacher Education Level for the teachers' database.

Logistic regressions [17] was the regression technique used both in the categorical attribute transformation and in the global binary classifier for its properties previously mentioned such as non-linear mapping capacity, ease of interpretation, ease of use, low data consumption and continuous output and also for its implementations freely available.

Considering the comparison assesses the difference in the discriminant power of the ranking score from all the algorithms, their performances are measured on each data fold by an area metric and a single point metric. The Area Under the Curve of the Receiver Operating Characteristics (AUC_ROC) [20] was calculated by just adding 0.5 to the Area Under the Curve of the Kolmogorov-Smirnov distribution [21]. The point metric was the Maximum Kolmogorov-Smirnov distance between the cumulative distribution functions of the target and complementary classes (Max_KS2) [22]. Both metrics used are invariant to score range and scale.

V. EXPERIMENTAL RESULTS AND INTERPRETATION

The results on Table I show that for both performance metrics the proposed approach (RGT) and the human expert weighted approach (WGT) are systematically better than the mode approach in all folds. However, RGT and WGT have similar performance in all folds.

TABLE I. PERFORMANCE OF THE BINARY CLASSIFIER FOR ALL APPROACHES MEASURED BY THE AUC_ROC AND MAX_KS2 ON EACH FOLD.

| Metrics | AUC_ROC | | | Max_KS2 | | |
|---|---|---|---|---|---|---|
| Fold No. | RGT | WGT | Mode | RGT | WGT | Mode |
| 1 | 0.891 | 0.887 | 0.832 | 0.615 | 0.640 | 0.582 |
| 2 | 0.889 | 0.880 | 0.834 | 0.607 | 0.610 | 0.585 |
| 3 | 0.863 | 0.853 | 0.796 | 0.552 | 0.547 | 0.520 |
| 4 | 0.846 | 0.852 | 0.773 | 0.523 | 0.544 | 0.457 |
| 5 | 0.876 | 0.885 | 0.841 | 0.610 | 0.629 | 0.599 |
| 6 | 0.869 | 0.868 | 0.814 | 0.585 | 0.560 | 0.530 |
| 7 | 0.878 | 0.870 | 0.842 | 0.612 | 0.602 | 0.626 |
| 8 | 0.890 | 0.884 | 0.818 | 0.642 | 0.615 | 0.558 |
| 9 | 0.844 | 0.841 | 0.766 | 0.542 | 0.536 | 0.446 |
| 10 | 0.864 | 0.869 | 0.813 | 0.599 | 0.586 | 0.509 |
| Mean | 0.871 | 0.869 | 0.813 | 0.589 | 0.587 | 0.541 |
| Median | 0.873 | 0.870 | 0.816 | 0.603 | 0.594 | 0.544 |
| Std.Dev. | 0.017 | 0.016 | 0.027 | 0.038 | 0.038 | 0.060 |

TABLE II. CLASSIFIER PERFORMANCE DIFFERENCES PAIRED T-TESTS BETWEEN ALL 3 APPROACHES FOR AUC_ROC AND MAX_KS2 METRICS AT 0.01 SIGNIFICANCE LEVEL.

| Metrics | Approaches | Mean | Std.Dev. | LimInf | LimSup | p-Val. |
|---|---|---|---|---|---|---|
| AUC_ROC | RGT-WGT | 0.0021 | 0.0067 | -0.0027 | 0.0069 | 0.348 |
| AUC_ROC | RGT-Mode | **0.0581** | **0.0148** | **0.0475** | **0.0687** | **0.000** |
| AUC_ROC | WGT-Mode | **0.0560** | **0.0150** | **0.0453** | **0.0667** | **0.000** |
| Max_KS2 | RGT-WGT | 0.0018 | 0.0185 | -0.0114 | 0.0150 | 0.766 |
| Max_KS2 | RGT-Mode | **0.0475** | **0.0367** | **0.0213** | **0.0737** | **0.003** |
| Max_KS2 | WGT-Mode | **0.0457** | **0.0350** | **0.0206** | **0.0708** | **0.003** |

The hypothesis tests (Table II, in boldface figures) show that both approaches that combine the influence of all categories in the attribute (histogram) are significantly different from the mode for granularity change in the binary classification problem presented.

VI. CONCLUSIONS

This paper has presented a systematic approach for producing granularity transformations of categorical attributes in binary decision problems. The concepts in the lower levels are summarized in a single continuous indicator learned from the attributes´ distribution histograms via regression on an independent sample, as a feature. It preserves the concept of the attribute in the original lower grain adding collective meaning to it, independent of any human knowledge about the application domain.

Thorough experimental procedure proved that the proposed granularity transformation produced statistically significant improvement in performance in the quality assessment of Brazilian secondary schools compared to the typical and simple distribution mode approach. It also showed that in the few cases where human knowledge is available, the proposed approach is equivalent to the existing WGT approach with human defined weights. Therefore, in the general case of having no assumption of human knowledge availability, the proposed approach presents a superior performance in binary decision problems.

Further experiments will be carried out on an even more realistic fashion by learning the distributions weights and developing the model from data from the previous education year and assessing the performance on the following year. There are several binary problems yet to be addressed in education at several stages of the process. There are also other domains with characteristics that suit the application of the proposed approach such as breast cancer diagnosis.